\documentclass[conference]{IEEEtran}
\IEEEoverridecommandlockouts
\usepackage{cite}
\usepackage{amsmath,amssymb,amsfonts}
\usepackage{algorithmic}
\usepackage{graphicx}
\usepackage{color}
\usepackage{textcomp}
\usepackage{fourier} 
\usepackage{array}
\usepackage{makecell}
\usepackage{flushend}
\usepackage{tikz}

\def\BibTeX{{\rm B\kern-.05em{\sc i\kern-.025em b}\kern-.08em
    T\kern-.1667em\lower.7ex\hbox{E}\kern-.125emX}}

\def\mycopyrightnotice{%
  \begin{minipage}{\textwidth}
  \small
  Copyright~\copyright~2019 IEEE. Personal use of this material is permitted. Permission from IEEE must be obtained for all other uses, in any current or future media, including reprinting/republishing this material for advertising or promotional purposes, creating new collective works, for resale or redistribution to servers or lists, or reuse of any copyrighted component of this work in other works by sending a request to pubs-permissions@ieee.org.
  
  \vspace{1.0cm}
  Accepted and published in: Proceedings of the 2022 17th International Conference on Control, Automation, Robotics and Vision (ICARCV), 2022, pp. 709-714, doi: https://doi.org/10.1109/ICARCV57592.2022.10004330.
  
  \vspace{1.0cm}
  This is the preprint only. The final version of the paper is available at: https://ieeexplore.ieee.org/document/10004330
  
  \vspace{1.0cm}
  Cite this paper as:
  
  \vspace{0.5cm}
  D. Belter, "Informed Guided Rapidly-Exploring Random Trees*-Connect for Path Planning of Walking Robots," Proceedings of the 17th International Conference on Control, Automation, Robotics and Vision (ICARCV), December 11-13, 2022, Singapore, s. 709-714, 2022, doi: https://doi.org/10.1109/ICARCV57592.2022.10004330
  \end{minipage}
}
\makeatother
    
\begin{document}
\mycopyrightnotice

\title{Informed Guided Rapidly-exploring Random Trees*-Connect for Path Planning of Walking Robots\\
\thanks{This research was supported by EU Horizon 2020 project THING.}
}

\author{Dominik Belter$^{1}$
\thanks{The work was supported by the National Science Centre, Poland, under research project no UMO-2019/35/D/ST6/03959.}
\thanks{$^{1}$Authors are with
          Institute of Robotics and Machine Intelligence,
          Poznan University of Technology,
          60-965 Pozna\'{n}, Poland
        {\tt\small dominik.belter@put.poznan.pl}}%
}

\maketitle

\begin{abstract}
In this paper, we deal with the problem of full-body path planning for walking robots. The state of walking robots is defined in multi-dimensional space. Path planning requires defining the path of the feet and the robot's body. Moreover, the planner should check multiple constraints like static stability, self-collisions, collisions with the terrain, and the legs' workspace. As a result, checking the feasibility of the potential path is time-consuming and influences the performance of a planning method. In this paper, we verify the feasibility of sampling-based planners in the path planning task of walking robots. We identify the strengths and weaknesses of the existing planners. Finally, we propose a new planning method that improves the performance of path planning of legged robots.
\end{abstract}

\section{Introduction}

Path planning methods define a sequence of robot states that connect the current and goal configurations of a robot. A basic benchmark for path planning methods is collision-free planning for a point-like object in the 2D environment. More challenging benchmarks take into account the 3D shape of the object and collisions with 3D obstacles. However, in both cases, constraints checking is fast and does not influence significantly the path planning algorithm. In contrast to simple path planning scenarios, a path planning problem for walking robots is defined in multi-dimensional space. The body of a walking robot has 6 degrees of freedom. Each leg can move independently in 3D space, so the path planner should determine the configuration of each leg, and contact with the ground. Taking into account 3 DoFs freedom for leg, six-legged and four-legged robots have 24 and 18 degrees of freedom, respectively. Moreover, the path planner should avoid many constraints like self-collision, collision with the terrain, stability, and workspace of the robot. As a result, determining a configuration of a robot and constraints checking takes a significant part of the path planning method.

Path planning methods can be divided into two main groups: grid-based and sampling-based. In the first group, the search space is divided into cells. Well-known methods like A* or D* can be applied to find the optimal path for the robot if the search space is low-dimensional and the number of cells is relatively small. In a multi-dimensional space, these methods do not work well due to the size of the obtained graph and require additional techniques to determine a feasible path for a robot~\cite{Xu2022}. In contrast to this approach, sampling-based methods operate in continuous search space and find the path for the samples obtained randomly from the robot's environment. In this group, we can find RRT-based planners like RRT and RRT-Connect~\cite{Kuffner2000} that are probabilistically complete but do not guarantee the optimal solution. The asymptomatically optimal and complete version of the sampling-based planner named RRT* was proposed in~\cite{Karaman2011}. The greedy version of this method that utilizes two trees was proposed later by Klemm {\em et al.}~\cite{Klemm2015} and also applied to the path planning of a six-legged walking robot. Finally, the method that utilizes informed sampling is proposed in~\cite{Gammell2014}. The greedy version named Informed RRT*-Connect algorithm is proposed in~\cite{Mashayekhi2020}. At the same time, hybrid approaches that combine grid and sampling-based methods have been developed to improve the results for multi-dimensional problems~\cite{Belter2019,Vonasek2018}.

\begin{figure}[!t!]
    \centering
    \includegraphics[width=0.9\columnwidth]{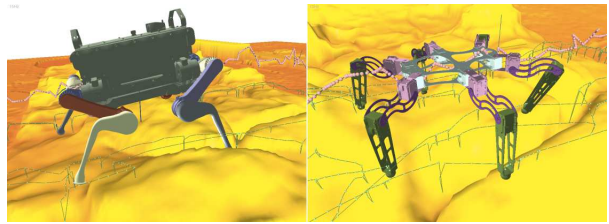}
    \caption{Path planning for legged robots: the planner simultaneously defines local body and feet motion (blue lines and pink dots) and a global path that avoids larger obstacles}
    \label{fig:intro}
    \vspace{-0.4cm}
\end{figure}

In this paper, we explore the optimal sampling-based planners in the task of path planning for multi-legged walking robots. We have implemented a set of methods for path planning of walking robots and identified the main advantages and drawbacks of existing methods. Finally, we propose an Informed Guided Rapidly-exploring Random Trees*-Connect (IGRSC) algorithm that improves the planning time and the length of the planned path compared to other sampling-based methods.

\begin{figure*}[!t!]
    \centering
    \includegraphics[width=1.0\textwidth]{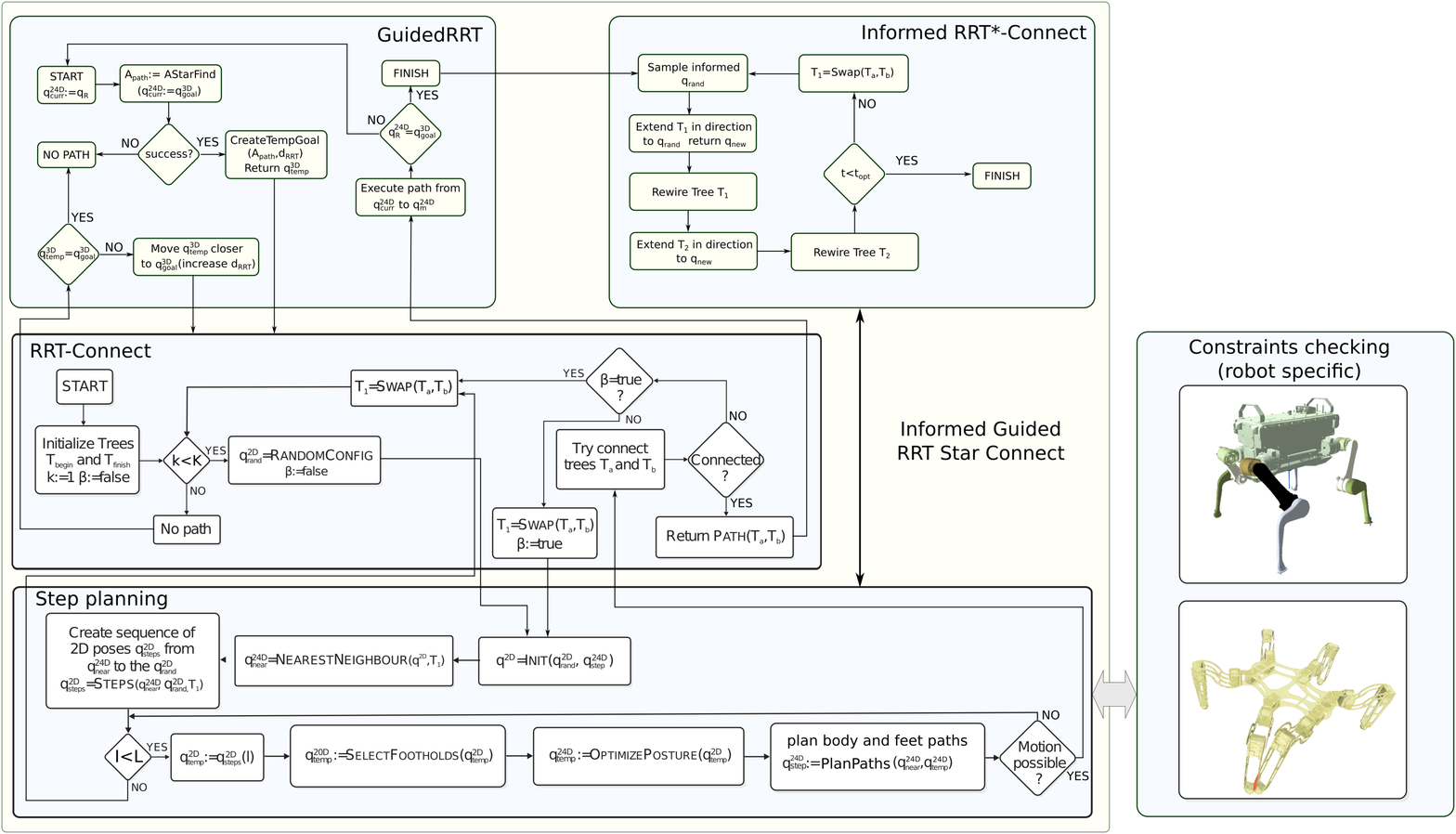}
    \caption{Architecture of the proposed Informed Guided RRT*-Connect (IGRSC) algorithm. The IGRSC utilizes the components of the state-of-the-art planner like RRT-Connect, GuidedRRT, and Informed RRT*-Connect during each stage of the planning}
    \label{fig:architecture}
\end{figure*}

\section{Related work}

Most of the planners dedicated to walking robots utilize elevation maps on the input. It comes from the fact that the main challenge of path planning for walking robots is related to rough terrain locomotion and the robot is higher than the terrain irregularities stored in the map. An example planner implemented on the ANYmal robot presented in~\cite{Fankhauser2018} finds footholds, optimizes the posture of the robot, and finds collision-free trajectories of the feet above the obstacles. The planner proposed in~\cite{Fankhauser2018} is local and the robot has to be guided by the operator to find the way around large obstacles. In~\cite{Belter2016jfr}, the RRT-based planner evaluates constraints and plans body and feet local paths but at the same time allows avoiding larger obstacles. This behavior is possible because, on top of the local motion planning methods, the path planning algorithm effectively evaluates search space to avoid obstacles. The foothold selection methods play an important role in the path planning for walking robots to guide the path of the robot through the regions with good footholds~\cite{fankhauser2018robust}. Tonneau2018 {\em et al.} compute a geometric approximation of the contact manifold that is used to find a feasible path of the walking robot using RRT planner~\cite{Tonneau2018}. The planner that takes into account the elevation map and assumes dynamic walking on rough terrain is presented in~\cite{Havoutis2013}. Also, the ANYmal robot is considered as a dynamic object and the trajectory of the robot is obtained by learning~\cite{Hwangbo2019}. Some research focuses on dynamic reflexes that stabilize the robot during walking~\cite{Focchi2019}. This approach allows finding a path for the robot's body and the online controller solves the problem of rough terrain locomotion. In this research, we assume that the robot is walking carefully on rough terrain thus we do not consider the dynamic properties of the robot.

Recently, planners using Convolutional Neural Networks like MPNet became popular~\cite{Qureshi2021}. The main advantage of neural-based approaches is the inference time that is significantly lower than for the typical approach for constraints checking~\cite{Belter2016jfr}. However, the MPNet has never been implemented to define the path or the full-body trajectory of a walking robot. Instead, neural networks are used to check constraints like the workspace of the legs and determine foothold~\cite{Villareal2019}. The learning-based approach that utilizes exteroceptive and proprioceptive perception on the ANYmal robot is presented in~\cite{Takahiro2022}. With this approach, the robot is capable of walking in challenging natural (mountains) and urban environments.

Recently, improving techniques for sampling-based planners have been proposed. In RRT* algorithm~\cite{Karaman2011} the neighbors are searched within a range that is reduced with the progress of the algorithm. Moreover, the rewiring procedure searches for a shorter connection in the tree. Both procedures make the RRT* asymptotically optimal. Gammel {\em et al.} improved the RRT* planner by sampling from ellipsoid when the initial path is found by the RRT* planner~\cite{Gammell2014}. Another "Informed" strategy has been implemented in methods that use a path found in a discretized space to guide a multi-dimensional search space using sampling-based methods~\cite{Belter2016jfr,Cizek2017,Vonasek2018}.  Even though the guided methods are not asymptotically optimal, they are proven to be efficient in multiple tasks. Because the RRT*-Connect has proven to be more efficient than the standard version~\cite{Klemm2015,Mashayekhi2020} we focus on the planners that simultaneously extend two trees starting in the initial and goal position of the robot. 

\subsection{Approach and Contribution}
\noindent 
In this work, we extend our previous approach to path planning of a six-legged robot~\cite{Belter2016jfr}. We utilize the foothold selection method proposed in~\cite{Belter2016jfr} and constraints checking procedures for self-collisions, static stability, and legs workspace proposed in~\cite{Belter2019}. First, we generalize and implement the RRT-based local motion planner for the quadruped walking robot ANYmal. Second, we propose a new sampling-based motion planning algorithm that effectively explores the search space and produces a kinematically plausible path for walking robots.

The main contributions of this paper include the following:
\begin{enumerate}
  \item Generalization of local path planner including posture optimization and body path planning for quadruped robots,
  \item A new planner named Informed Guided Rapidly-exploring Random Trees*-Connect (IGRSC) that uses multi-level planning for fast-path finding and optimization,
  \item Implementation of the set of planning methods and verification on the models of six and four-legged robots.
\end{enumerate}

\section{IGRSC Path Planner}
\label{system}

\subsection{Global planning methods}

The architecture of the proposed method is presented in Fig.~\ref{fig:architecture}. We utilize two meta-planners on the top of the proposed planner: GuidedRRT~\cite{Belter2016jfr} and Informed RRT*-Connect~\cite{Mashayekhi2020}. In the first stage, we utilize the GuidedRRT planner that effectively guides the local RRT-Connect planner~\cite{Kuffner2000} using A* planner. A* planer searches for the path on the map with reduced resolution and simplified cost prediction~\cite{Belter2016jfr}. Then, we define a temporary goal that is $d_{\rm RRT}$ far from the current position of the robot. The path between the current robot position and the temporary goal is found using the RRT-Connect planner that effectively works with our local full-body step planner~\cite{Belter2019}. If the RRT-Connect successfully finds the path, we change the current position of the robot ${\bf q}_{\rm curr}^{\rm 24D}$ along the path found by the RRT-Connect (``execute path'' step in Fig.~\ref{fig:architecture}). If RRT-Connect fails, we move the temporary goal in the direction to the goal configuration. In the worst-case scenario, the RRT-Connect searches for the path between the initial configuration of the robot ${\bf q}_{\rm R}$ and the goal configuration ${\bf q}_{\rm goal}^{\rm 3D}$. As a result, our algorithm inherits the properties of the RRT-Connect planner and is probabilistically complete. 

In the second stage, we utilize the Informed version of the RRT*-Connect algorithm~\cite{Mashayekhi2020}. To this end, we convert the path found by the GuidedRRT algorithm into the tree and inject this solution into the RRT*-Connect algorithm. As a result, we skip the first stage of the RRT*-Connect planer and we run only the optimization stage that samples the search space using the ellipsoid created from the injected path. We found that this strategy is more efficient in the path planning problem of walking robots than running RRT*-Connect from the beginning. The GuidedRRT is more efficient in finding the initial path than a random sampling of the search space in the RRT*-Connect algorithm.

The Informed RRT*-Connect algorithm presented in Fig.~\ref{fig:architecture} alternately extends two trees. First, the tree is extended in the direction given by the sample obtained from the ellipsoid that overlaps the current path. We compute the ellipsoid using the start and goal positions of the robot and the length of the initial path like in~\cite{Gammell2014}. Then, we plan a sequence of steps that move the robot in the direction to the random node. After each successful tree extension operation, we check if the trees can be connected by planning the full-body path of the robot between the new node $q_{\rm new}$ and the neighboring node from the second tree. In the next step, we run the procedure (rewire tree) that check if we can connect existing nodes in the tree with the new node $q_{\rm new}$ and shorten the path to the root node in the tree. In this procedure, we compute the sphere with a gradually decreasing radius to reduce the number of nodes that are checked by the procedure. In the application of a walking robot, checking all possible shortcuts in the tree is computationally expensive because it requires full-body path planning. Thus, in our implementation, we reduce the number of potential nodes to $N$ nearest neighbors found in the sphere (\ref{radius}) ($N$=3 in the presented experiments). If the shortcut is found, we add the new branches to the tree and remove the branches that create loops in the graph. In the next iteration of the algorithm, the second tree is extended. 

In the proposed motion planning algorithm we significantly reduce the number of nodes in the shortcut searching stage. However, the Informed RRT*-Connect algorithm is asymptotically optimal if we keep the relations (1) and (2):

\begin{equation}
\gamma_{\rm RRT^{*}} > (2 \cdot (1+\frac{1}{d}))^{\frac{1}{d}} \cdot (\frac{\mu(X_{free})}{\zeta_d})^{\frac{1}{d}},
\end{equation}

\begin{equation}
\label{radius}
r(n) = \gamma_{\rm RRT^{*}}(log(n)/n)^{\frac{1}{d}},
\end{equation}

\noindent where $r(n)$ is the radius of the sphere, $n$ is the number of samples, $d$ is the dimension of the search space, $\mu(X_{free})$ is the Lebesgue measure of the free space, and $\zeta_d$ is the volume of the unit ball in the $d$-dimensional Euclidean space. However, the goal of using the RRT*-Connect algorithm is to optimize the initial path found by the GuidedRRT, and with the reduced number of checked connections in the tree ($r(n)$ is small) and without the optimality of the RRT* it can still effectively reduce the length of the initial path.

\subsection{Local path planner}

The main task of the local path planner is to define a single step of the robot in the direction given by the higher-level planners (RRT-Connect or Informed RRT*-Connect). RRT-Connect uses a local path planner to plan a single step during the {\em extend} procedure while the RRT*-Connect plans a sequence of steps in the direction to the goal pose. We adopted the planner implemented for the six-legged walking robot Messor II~\cite{Belter2016jfr}. We generalize the approach presented in~\cite{Belter2016jfr} to use the local planner also on four-legged robots. Some of the techniques are the same for the six-legged and four-legged robots. However, planning the motion of the four-legged robot is more challenging due to the much smaller range of static stability regions. For the six-legged robots, we accepted sudden changes in the robot's body direction. In this paper, we focused to guarantee the smoothness of the trajectory since this aspect plays an important role during body balancing in quadruped locomotion. During the step planning procedure, we check constraints like self-collisions $C_{\rm free}$, workspace $C_{\rm work}$, kinematic margin $d^{\rm KM}$, collisions with the terrain $C_{\rm free}$, and static stability $C_{\rm free}$ employing Gaussian Mixtures to speed up computations~\cite{Belter2019}.

The local path planning starts with defining a set of 2D positions of the robot in the direction to the goal node $q_{\rm rand}^{\rm 2D}$. Starting from the longest step, we plan the robot's motion and stop the procedure when a plausible path is found. The procedure starts with the foothold selection method~\cite{Belter2011}. Both robots are equipped with ball-like feet and the feet sizes are similar. Thus, we use the same strategy for both robots~\cite{Belter2019ICRA}. Next, for the given positions of the feet, the robot optimizes its posture. The goal of this procedure is to keep the robot stable and preserve the high kinematic range of motion. Thus, in the original version designed for the six-legged robot, we optimized (maximized) kinematic margin $d^{\rm KM}$~\cite{Belter2019}. This procedure applied on the four-legged robot ANYmal returns the pose of the body that is very low above the ground. In this case, the kinematic margin is maximal but the robot is close to the terrain, and the energy consumption is higher due to higher joint torques. Thus, in the proposed procedure, we added the offset along the $z$ axis $p_z$ to the cost function to keep the robot high above the ground: 

\begin{align}
\label{postureOpt}
    \operatorname*{argmax}_{{\mathbf p}_R} (d^{\rm KM}+p_z)({\mathbf p}_R \in C_{\rm free} \cap C_{\rm stab} \cap C_{\rm work},
\end{align}

\noindent where ${\mathbf p}_R$ is the posture of the robot defined by the distance to the ground $p_z$ and the robot's body inclination~\cite{Belter2019}.

In the next step, the procedure plans the path between the initial and optimized posture of the robot for each foot and the robot's body~\cite{Belter2019}. The initial path is planned with the assumption that the motion is linear. This assumption does not guarantee that the robot will remain stable and collision-free between the initial and goal pose of the robot so for each point of the planned linear path we optimize the position of the swing legs and move the robot's body to the stable position. As a result, the robot balances sideways during the execution of the planned path to preserve stability. This behavior is rarely observed for a six-legged robot but is very common for a quadruped robot. To prevent rough changes of the robot's body we propose to use B-spline interpolation defined in SE3 that guarantees a smooth and differentiable path in 3D space~\cite{Nowicki2020,Sommer2020}. To this end, we iterate along the planned path and search for the largest displacement of the robot's body that stabilizes the robot ${\bf q}_{\rm stab}^{\rm step}$. Then we compute the B-spline for three knots ${\bf p}_i$: initial pose ${\bf q}_{\rm init}^{\rm step}$, stabilized pose ${\bf q}_{\rm stab}^{\rm step}$, and goal pose ${\bf q}_{\rm goal}^{\rm step}$. The procedure returns a sequence of smooth SE3 poses in-between the defined nodes. The B-spline is defined by the knots ${\bf p}_i$, times $t_i=t_0+i\bigtriangleup t$, and a set of spline coefficients $B_{i,k}(t)$:

\begin{align}
\label{bcoefs}
    {\bf p}(t)=\sum_{i=0}^{N} B_{\rm i,k} {\bf p}_i,
\end{align}

\noindent the coefficients are computed using De Boor-Cox recurrence formula~\cite{Sommer2020}:

\begin{align}
\label{bcoefs1}
    B_{\rm i,0}(t) =\begin{cases} 1 & for~t_i\leq t < t_{\rm i+1} \\
                     0 &  otherwise
       \end{cases},
\end{align}

\begin{align}
\label{bcoefs2}
    B_{\rm i,j}(t) = \frac{t-t_i}{j\bigtriangleup t} B_{i,j-1}(t)+\frac{t_{\rm i+j+1}-t}{j\bigtriangleup t} B_{i+1,j-1}(t).
\end{align}

The obtained path is differentiable and guarantees quasi-static stability.

\section{Results}

\begin{figure}[!t!]
    \centering
    \includegraphics[width=1.0\columnwidth]{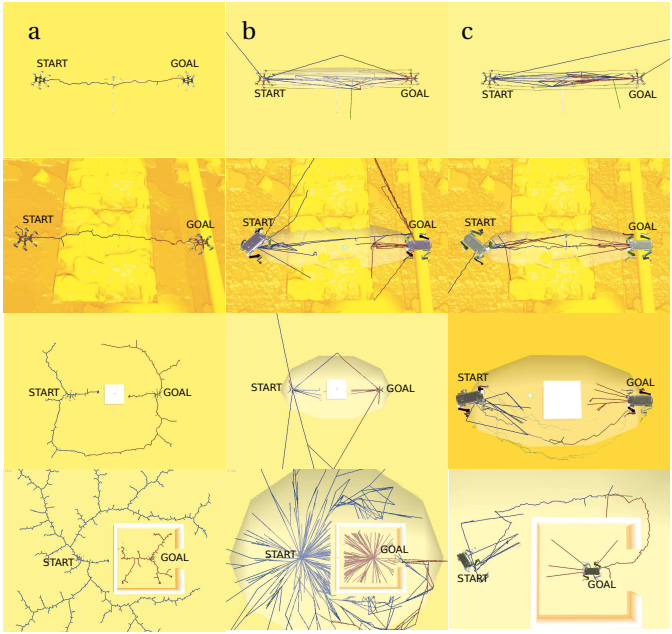}
    \put(-243,224){a} \put(-162,224){b} \put(-79,224){c}
    \caption{Comparison between exploration strategies for the RRT-Connect (a), RRT*-Connect (b), and IGRSC (c) on various maps. The white ellipsoids represent the current sampling region for the informed algorithms. Note that the proposed planner is general and the implementation is the same for the six-legged and quadruped robots. The resulting trees does not depend significantly on the robot type (kinematic)}
    \label{fig:results_tree}
\end{figure}

We have verified the proposed method and compared it to other planners in a systematic way on a set of motion planning challenges. In all experiments, the goal position of the robot is 4.6~m in from the initial position of the robot. In the first scenario, the robot walks on flat terrain without additional obstacles. In this experiment, we check which planner is the fastest in a simple planning task where the local path planning plays an important role. In the second scenario, we verify the planner on the map of the irregular terrain. The highest point of the map, that the robots should walk on is higher than the Messor II robot.  The third environment contains a box in the center of the map. Even though this environment is simple, it causes problems for sampling-based planners. First, the ``Connect'' version of the RRT planners extends in the direction to the second tree during each iteration. At the beginning of planning this direction is occupied by the obstacle and the extension of the tree fails very often. Second, the ``informed'' version of the planner creates an ellipsoid in the point between the start and goal position. This position is used to generate random samples and most of them are located under the obstacle in the center of the map. As a result, the planning time for these methods is longer. In the last scenario, we use the ``bug trap'' environment to verify the planners in the challenging environment where the top-level techniques play the most important role. The entrance to the trap is 1.2~m wide for the Messor II robot and 1.7~m for the ANYmal robot.

We have compared five various sampling-based algorithms: RRT-Connect~\cite{Kuffner2000}, GuidedRRT~\cite{Belter2016jfr}, RRT*-Connect~\cite{Klemm2015}, Informed RRT*-Connect~\cite{Mashayekhi2020}, and the proposed IGRSC planner. The Informed RRT*-Connect method uses the RRT*-Connect algorithm to find the feasible path. Then, the informed sampling is used to optimize the path for 180~s. Similarly, IGRSC optimizes the obtained trajectory from GuidedRRT for 180~s. We register two metrics: the average planning time $t_p$ and the average length of path $\overline{p}$ found by the planner. The statistics are obtained for 10 trials on each map using each verified algorithm. The experiments are performed on the computer with the i9-9900KF processor.

\begin{figure}[!t!]
    \centering
    \includegraphics[width=1.0\columnwidth]{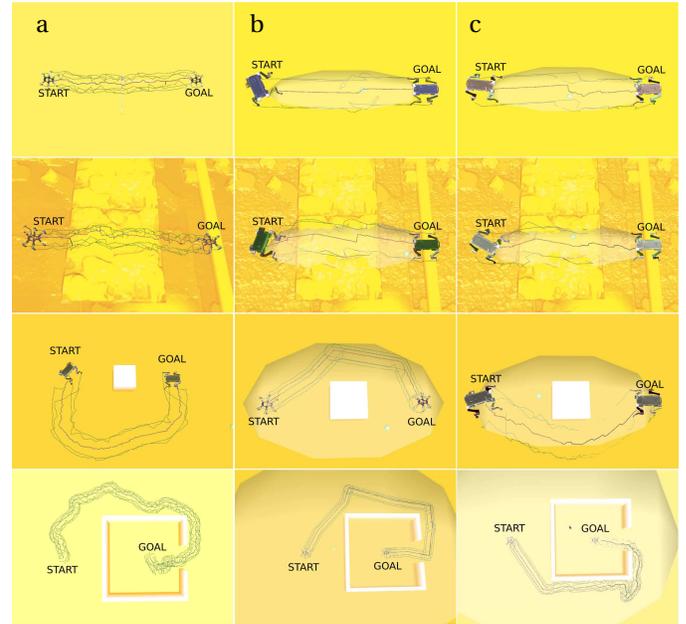}
    \put(-243,224){a} \put(-162,224){b} \put(-79,224){c}
    \caption{Example paths found by RRT-Connect (a), RRT*-Connect (b), and IGRSC (c)}
    \label{fig:results_path}
\end{figure}

In Fig.~\ref{fig:results_tree}, we demonstrate example trees generated by verified motion planning algorithms. The screenshots are created at the end of planning and illustrate the different strategies used by motion planning methods. The branches in the Informed RRT*-Connect (Fig.~\ref{fig:results_tree}a) are straight because the planner plans a sequence of steps, while RRT-Connect (Fig.~\ref{fig:results_tree}a) plans a single step during the tree extension. The IGRSC algorithm (Fig.~\ref{fig:results_tree}c) combines a path found by the guidedRRT that is generated by the RRT-Connect and straight lines generated by the Informed RRT*-Connect. The example paths found by the algorithms are presented in Fig.~\ref{fig:results_path}. Note that the RRT-Connect does not generate a feasible path in the limited time (1e4~s).

\begin{table*}[t]
\caption{Success rate (succ), average planning time $\overline{t}_p$ and obtained path length $\overline{p}$ on the Messor II robot. The line splits asymptomatically complete  methods and methods that optimize the path during the second stage of planning}
\vspace{-0.5cm}
\newcommand{\mc}[3]{\multicolumn{#1}{#2}{#3}}
\label{tab:results_messor}
\begin{center}
\setlength\tabcolsep{2.0pt}
\scriptsize
\begin{tabular}{l|ccccc|ccccc|ccccc|ccccc}
& \mc{4}{c}{flat} & \mc{4}{c}{rough terrain} & \mc{4}{c}{box} & \mc{4}{c}{bug trap}\\ 
 & $succ$ &  $\overline{t}_p$~[s] & $\sigma_t$~[s] & $\overline{p}$~[m] & $\sigma_{p}$~[m] & $succ$ &  $\overline{t}_p$~[s] & $\sigma_t$~[s] & $\overline{p}$~[m] & $\sigma_{p}$~[m] & $succ$ &  $\overline{t}_p$~[s] & $\sigma_t$~[s] & $\overline{p}$~[m] & $\sigma_{p}$~[m] & $succ$ &  $\overline{t}_p$~[s] & $\sigma_t$~[s] & $\overline{p}$~[m] & $\sigma_{p}$~[m]\\
RRT-Connect~\cite{Kuffner2000} & 1.0 & {\bf 3.4} & {\bf 0.3} & 5.01 & 0.10 & 1.0 & {\bf 7.0} & {\bf 2.3} & 5.14 & 0.11 & 1.0 & 90.5 & 19.2 & 10.45 & 1.79 & 0.2 & 443.5& 83.9& 17.64& 1.04 \\ 
GuidedRRT~\cite{Belter2016jfr} & 1.0 & 5.3 & 0.2 & 4.91 & 0.08 & 1.0 & 8.6 & 1.5 & 5.09 & 0.10 & 1.0 & {\bf 7.2} & {\bf 2.0} & 5.62 & 0.11 & 1.0 & {\bf 110.9} & {\bf 160.8} & 14.45 & 0.66\\ \hline
RRT*-Connect~\cite{Klemm2015} & 1.0 & 187.0 & 3.1 & 4.99 & 0.49 & 1.0 & 285.8 & 64.2 & 6.04 & 2.20 & 1.0 & 194.2 & 8.2 & 6.26 & 0.69 & 0.9 & 422.4 & 103.3& 15.20 & 0.52 \\ 
IRRT*-Connect~\cite{Mashayekhi2020} & 1.0 & {\bf 186.2} & {\bf 1.2} & {\bf 4.61} & {\bf 0.01} & 1.0 & 308.79 & 232.3 & 5.21 & 1.60 & 1.0 & 188.9 & 4.7 & 5.38 & 0.07 & 1.0 & 340.2& 100.9& 15.92& 1.97\\ 
IGRSC (our) & 1.0 & {\bf 187.2} & {\bf 0.4} & {\bf 4.61} & {\bf 0.02} & 1.0 & {\bf 196.9} & {\bf 9.8} & {\bf 4.73} & {\bf 0.05} & 1.0 & {\bf 188.0} & {\bf 1.2} & {\bf 5.35} & {\bf 0.12} & 1.0 & {\bf 233.8} & {\bf 20.0} & {\bf 13.86} & {\bf 0.50}\\ 
\end{tabular}
\end{center}
\vspace{-0.5cm}
\end{table*}

\begin{table*}[t]
\caption{Success rate (succ), average planning time $\overline{t}_p$ and obtained path length $\overline{p}$ on the ANYmal robot. The line splits asymptomatically complete  methods and methods that optimize the path during the second stage of planning}
\vspace{-0.5cm}
\newcommand{\mc}[3]{\multicolumn{#1}{#2}{#3}}
\label{tab:results_anymal}
\begin{center}
\setlength\tabcolsep{2.0pt}
\scriptsize
\begin{tabular}{l|ccccc|ccccc|ccccc|ccccc}
& \mc{4}{c}{flat} & \mc{4}{c}{rough terrain} & \mc{4}{c}{box} & \mc{4}{c}{bug trap}\\ 
 & $succ$ & $\overline{t}_p$~[s] & $\sigma_t$~[s] & $\overline{p}$~[m] & $\sigma_{p}$~[m] & $succ$ & $\overline{t}_p$~[s] & $\sigma_t$~[s] & $\overline{p}$~[m] & $\sigma_{p}$~[m] & $succ$ & $\overline{t}_p$~[s] & $\sigma_t$~[s] & $\overline{p}$~[m] & $\sigma_{p}$~[m] & $succ$ & $\overline{t}_p$~[s] & $\sigma_t$~[s] & $\overline{p}$~[m] & $\sigma_{p}$~[m]\\
RRT-Connect~\cite{Kuffner2000} & 1.0 & {\bf 3.1} & {\bf 0.2} & 5.06 & 0.06 & 1.0& 10.4 & 13.5 & 5.34 & 0.59 & 1.0 & 87.4 & 19.2 & 10.19 & 1.65 & 0.5 & 478.7 & 88.0 & 19.82 & 1.22 \\ 
GuidedRRT~\cite{Belter2016jfr} & 1.0 & 5.6 & 0.2 & 5.01 & 0.09 & 1.0 & {\bf 10.8} & {\bf 3.1} & 5.12 & 0.20 & 1.0 & {\bf 13.9} & {\bf 12.0} & 5.97 & 0.31 & 0.6 & {\bf 199.6} & {\bf 166.1} & 16.87 & 0.51\\ \hline
RRT*-Connect~\cite{Klemm2015} & 1.0 & 240.6 & 156.9 & 4.86 & 0.22 & 1.0 & 227.4 & 68.6 & 5.32 & 0.58 & 0.8 & 281.3 & 99.7 & 6.82 & 1.20 & 0.2 & 606.7 & 14.0 & 16.15 & 0.85 \\ 
IRRT*-Connect~\cite{Mashayekhi2020} & 1.0 & 199.4 & 26.5 & 4.66 & 0.07 & 1.0 & 264.0 & 91.9 & 4.79 & 0.06 & 1.0 & 200.2 & 8.0 & 6.96 & 2.47 & 0.1 & 500.3 & 0.0 & 16.51 & 0.00\\ 
IGRSC (our) & 1.0 & {\bf 187.0} & {\bf 0.4} & {\bf 4.63} & {\bf 0.02} & 1.0 & {\bf 189.8} & {\bf 1.7} & {\bf 4.74} & {\bf 0.05} & 1.0 & {\bf 196.2} & {\bf 14.3} & {\bf 5.65} & {\bf 0.16} & 0.6 & {\bf 321.7} & {\bf 63.6} & {\bf 16.01} & {\bf 1.12}\\ 
\end{tabular}
\end{center}
\vspace{-0.6cm}
\end{table*}


The results obtained for the six-legged Messor II robot are presented in Tab.~\ref{tab:results_messor}. In all experiments, the shortest average path was found by the proposed IGRSC algorithm. Only the Informed RRT*-Connect algorithm returns similar results on the rough terrain map. It is not straightforward to compare planning time because the optimization planning time for the informed version of the algorithms is set to a constant value. However, we can conclude that the RRT-Connect is efficient in tasks where the problem is solved locally (body path planning, feet trajectory planning) and does not require avoiding larger obstacles (experiment on flat and rough terrain). However, this strategy fails on the ``bug trap'' map. The RRT-Connect was unable to find the path in the limited-time set to 1e4~s. Moreover, on the flat and box maps the RRT-Connect has a shorter path length than RRT*-connect. This result suggests that planning a single step implemented in the RRT-Connect is a more efficient strategy than planning a sequence of steps in the random direction used in the RRT* planner. When the robot should take a detour to reach the goal position, the GuidedRRT and RRT* perform the best. On three maps the GuidedRRT finds the path faster than the RRT*. This result justifies our decision to use this algorithm to find the initial path using the GuidedRRT in the proposed IGRSC planner. 

The general conclusions that we draw from the experiments on the ANYmal robot (Tab.~\ref{tab:results_anymal}) are similar to those obtained on the Messor II robot. The RRT-Connect is efficient in finding the path in a simple environment (flat, rough terrain), but it performs worse or fails in environments where the robot has to take longer detours from the straight path. At the same time, GuidedRRT quickly finds the initial path. For the simpler maps, the Informed RRT* Connect and IGRSC returns similar and the shortest paths. In more demanding environments the IGRS works much better. The RRT*-Connect and Informed RRT*-Connect provided one path out of 10 trials in the limited time. 

\section{Conclusions}

In this paper, we propose a new path planning method for walking robots. We utilize the Guided RRT that uses the A* planner in the 2D search space to guide the RRT-Connect planner to find the path between the initial and goal position of the robot. To generalize and implement the planner on quadruped robots, we propose an improved strategy for posture optimization and B-spline-based path planning in SE3. Our experiments show that this method quickly finds a feasible path in various environments. Then, we optimize the path using the Informed RRT*-Connect strategy that simplifies the initial full-body path of the robot found by the GuidedRRT. In all experiments performed in this paper, the proposed IGRSC planner was capable to find the shortest path. Taking into account the time required for finding the plausible path and path optimization the proposed algorithm performs better in challenging environments, where the differences between planners are better visible.

The proposed method has some drawbacks. We utilize the greedy variant of the sampling-based planners (Connect), so we observe in some situations that the algorithm unsuccessfully tries multiple times to connect both trees through the obstacles. We believe that training a neural network to predict the locomotion cost in the neighbor of the current robot pose, would be used to guide the sampling strategy of the planner and improve significantly the planning time. In the future, we are also going to use the obtained near-optimal paths to train the MPNet-based neural planner.





\bibliographystyle{IEEEtran}
\bibliography{IEEEabrv,IEEEexample}

\end{document}